\documentclass[conference]{IEEEtran}
\IEEEoverridecommandlockouts
\usepackage{amsmath,amssymb,amsfonts}
\usepackage{algorithmic}
\usepackage{graphicx}
\usepackage{textcomp}
\usepackage{bm, bbm}
\usepackage{xcolor}
\usepackage{cite}
\usepackage{multirow}
\def\BibTeX{{\rm B\kern-.05em{\sc i\kern-.025em b}\kern-.08em
    T\kern-.1667em\lower.7ex\hbox{E}\kern-.125emX}}
\usepackage[accsupp]{axessibility}
\usepackage{url}

\usepackage{fancyhdr}
\begin{document}

\title{Path Generation and Evaluation in Video Games:\\A Nonparametric Statistical Approach}

\author{
    \IEEEauthorblockN{
    Daniel Campa\textsuperscript{1},
    Mehdi Saeedi\textsuperscript{2}, 
    Ian Colbert\textsuperscript{2}, 
    Srinjoy Das\textsuperscript{1} \thanks{{Correspondence to}: {srinjoy.das@mail.wvu.edu}, {icolbert@amd.com}}}
    \vspace{0.5em}
    \IEEEauthorblockA{\textsuperscript{1}School of Mathematical and Data Sciences, West Virginia University}
    \IEEEauthorblockA{\textsuperscript{2}AMD}
}
\IEEEaftertitletext{\vskip -0.5cm}
\maketitle

\begin{abstract}
Navigation path traces play a crucial role in video game design, serving as a vital resource for both enhancing player engagement and fine-tuning non-playable character behavior. Generating such paths with human-like realism can enrich the overall gaming experience, and evaluating path traces can provide game designers insights into player interactions.
Despite the impressive recent advancements in deep learning-based generative modeling, the video game industry hesitates to adopt such models for path generation, often citing their complex training requirements and interpretability challenges. To address these problems, we propose a novel path generation and evaluation approach that is grounded in principled nonparametric statistics and provides precise control while offering interpretable insights. Our path generation method fuses two statistical techniques: (1) nonparametric model-free transformations that capture statistical characteristics of path traces through time; and (2) copula models that capture statistical dependencies in space. For path evaluation, we adapt a nonparametric three-sample hypothesis test designed to determine if the generated paths are overfit (mimicking the original data too closely) or underfit (diverging too far from it). 
We demonstrate the precision and reliability of our proposed methods with empirical analysis on two existing gaming benchmarks to showcase controlled generation of diverse navigation paths. Notably, our novel path generator can be fine-tuned with user controllable parameters to create navigation paths that exhibit varying levels of human-likeness in contrast to those produced by neural network-based agents.
The code is available at \texttt{\url{https://github.com/daniel-campa/mf-copula}}.
\end{abstract}


\section{Introduction}
\label{intro}

Navigation path traces are essential for game designers, providing key opportunities to enhance the gaming experience. By evaluating these traces, designers can gain valuable insights into player interactions, enabling them to refine non-player character (NPC) behavior and optimize gameplay \cite{karpov2012believable, campbell2015clustering}. Additionally, generating novel paths to guide NPC behavior increases player engagement, especially when these paths emulate human-like navigation \cite{soni2008bots, pearce2022counter, milani2023navigates}. Recent advancements in data-driven path generation and evaluation largely rely on deep learning. Data-driven path generation is commonly done by training navigation agents with reinforcement learning (RL) algorithms, which have achieved significant success by learning complex policies through interactions with dynamic environments \cite{alonso2020deep, team2021open}. For path evaluation, neural networks offer impressive precision in distinguishing artificial from human game traces, providing nuanced assessments of gaming authenticity \cite{devlin2021navigationturingtestntt}. Despite these advances, the video game industry has been slow to adopt deep learning models, citing challenges such as complex training, high retraining costs, and limited interpretability, which hinder practical development, maintenance, and optimization \cite{jacob2020itsunwieldytakeslot, aytemiz2021acting}.

\begin{figure}[t]
    \centerline{\includegraphics[width=0.5\textwidth]{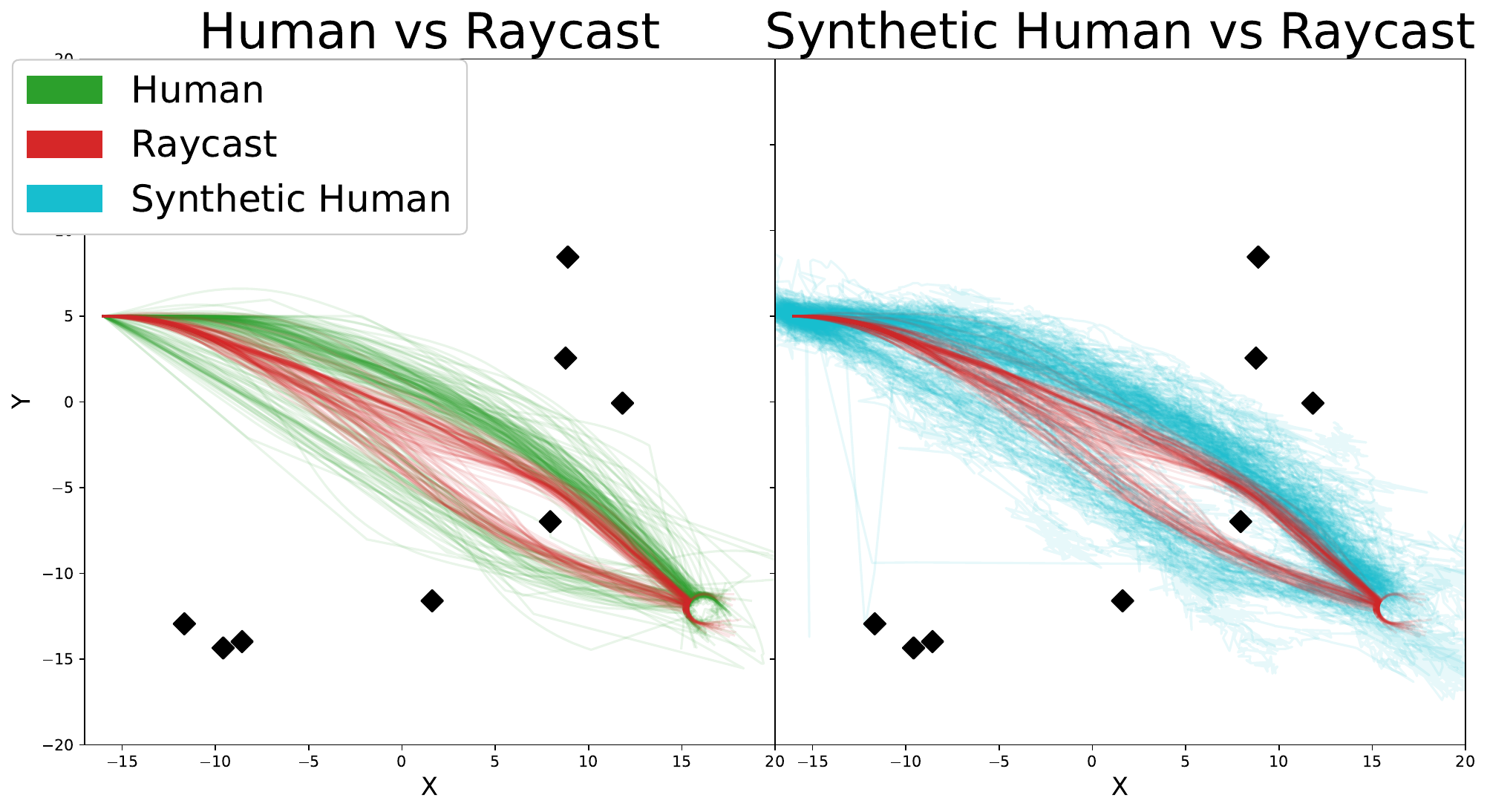}}
    \caption{The original data consists of human (green) and raycast-based RL agents (red) navigating through a 2D obstacle course, with obstacles shown as black diamonds \cite{10.3389/fpsyg.2021.725932}. Our synthetically generated human trajectories (blue) are visually more human-like than the raycast agents. We analytically show this in Section~\ref{ev} using statistical hypothesis testing.}
    \label{fig:frontiers-experiments}
    \vspace{-0.5cm}
\end{figure}

In contrast to deep learning based methods, statistical approaches can work with limited amounts of data for training, and offer intuitive insights that can translate to tangible actions.
Accordingly, we propose a nonparametric statistical approach for generating synthetic path traces. Unlike model-based approaches such as ARMA \cite{politis2019time} and GARCH \cite{francq2019garch}, our proposed nonparametric framework captures the temporal characteristics of the data over individual dimensions of the paths without relying on predefined equations or distributional assumptions. 
Using this in conjunction with copulas \cite{nelsen2006introduction}, our proposed approach ensures flexibility and robustness in generating synthetic paths while capturing time-varying dependencies across dimensions. Our method is particularly suited for video game path generation, where the data is often multivariate with a relatively low dimensionality (typically $\leq$3 dimensions). In such cases, individual path dimensions can have varying degrees of non-stationarity and complex spatiotemporal dependencies across dimensions, both of which are captured by our copula-based modeling approach. Examples of synthetic paths generated by our algorithm are shown in Fig. \ref{fig:frontiers-experiments}.

% \newpage

Our methodology caters to game analytics in two key directions.
First, our data augmentation proposal supports game development by providing developers with both generation and evaluation of large-scale path data. 
Second, by learning from real player data, our method yields synthetic trajectories that incorporate the nuanced decision patterns of human players or advanced NPCs. Although left for future work, such patterns can then be used to guide NPC movement strategies, enabling behaviors that feel more natural and player-like, rather than purely algorithmic or predictable \cite{yannakakis2018artificial}. This is particularly useful for companion AI, cooperative NPCs, or enemy behaviors aiming to emulate a human opponent \cite{panwar2022npcaiuscase}.
Our generator’s data and time efficiency is especially valuable here, as acquiring extensive real-world human data often proves infeasible \cite{el2013game}.

Within this context, the contributions in this paper are four-fold: (1) We apply a model-free transformation on each component of the multivariate path time series to capture time-varying non-stationarity in each dimension; (2) We construct a novel framework for generating synthetic paths by linking the residuals generated from the model-free transformation in each dimension with a time-varying Gaussian copula; (3) We adapt a nonparametric hypothesis test to verify whether the synthetically generated  paths are essentially indistinguishable or overly different from the original training data; and (4) We demonstrate the effectiveness of our proposed model-free-copula technique and our time series adapted hypothesis test on two datasets from gaming applications.

\section{Background \& Related Work}
\label{bg}

Our work uses copulas with a nonparametric model-free approach for synthetic data generation. In addition, we use a nonparametric hypothesis testing framework for determining the validity of the synthetic time series generated by our copula-based algorithm. This section reviews previous work in this domain as a motivation for our algorithmic innovations.

\textbf{Deep Generative Models (DGMs).} Variational autoencoders (VAEs) \cite{kingma2013auto}, generative adversarial networks \cite{goodfellow2014generative}, normalizing flows \cite{papamakarios2021normalizing} and diffusion models \cite{ho2020denoising} have been used extensively for data augmentation in several applications such as image classification \cite{antoniou2017data}, text classification \cite{ghadekar2023text} and acoustic modeling \cite{nishizaki2017data}, among others. DGMs have also been used for generating synthetic paths for RL applications in finance \cite{liu2022synthetic}, proprioceptive control tasks \cite{lu2023synthetic}, and even goal-oriented navigation \cite{jain2023learning}. However, generating synthetic paths using DGMs typically requires large amounts of training data to effectively capture underlying distributions, and retraining for new tasks is often prohibitively expensive. Moreover, DGMs trained with insufficient amount of samples are highly likely to underfit and generalize poorly\cite{lu2023machine}. 

\textbf{Copulas.} As an alternative, statistical methods like copulas in conjunction with univariate time series models  offer a promising solution for synthetic generation by flexibly modeling the spatiotemporal dependencies between variables, even with smaller datasets. Copulas can effectively capture the joint distribution of multi-dimensional RL paths as they evolve in time,
making them a practical choice for generating synthetic paths in scenarios where DGMs face limitations \cite{nelsen2006introduction}. 
In previous research for applications such as financial forecasting, energy price prediction and wind speed modeling, copulas have been used with parametric time series models such as ARMA and GARCH to capture dependencies between variables \cite{jondeau2006copula,li2013copula, gregoire2008using}. For instance, Copula-GARCH models use GARCH processes to handle volatility and apply copulas to capture non-linear dependencies across residuals from different series, enabling the modeling of financial risks more accurately\cite{jondeau2006copula}. Similarly, in ARMA-based models, copulas offer a way to extend beyond univariate forecasting by linking residuals between multiple time series, thereby capturing complex dependencies more flexibly \cite{li2013copula}. 
 These parametric approaches, however, come with inherent assumptions which may limit their adaptability to real-world datasets. For example, GARCH assumes specific distributions for error terms and focuses on heteroskedasticity\footnote{Heteroskedasticity means that the variance of errors is not constant across the observations.}, while ARMA relies on linear additive models and stationarity assumptions. Such strict assumptions can lead to model misspecification when data exhibit higher order non-stationarity or non-linear dependencies, limiting the robustness of synthetic data generation in practical applications. This is particularly problematic in RL, where generating realistic paths requires capturing intricate dependencies without overly simplifying assumptions \cite{liu2022synthetic}.

\textbf{Hypothesis Testing.} A key consideration for any synthetic data framework is to determine the validity of the generated samples. Standard approaches to evaluating navigation paths rely on expert human judges that require time-intensive manual effort \cite{milani2023navigates}, or coarse metrics that fail to capture fine-grained details \cite{karpov2012believable, kirby2009companion}, or domain-specific automated proxies based on machine learning models \cite{devlin2021navigationturingtestntt} which need training and extensive fine-tuning. However, a more principled and interpretable approach is to apply hypothesis tests to objectively determine the quality of the synthetic data. Two-sample tests that work with high-dimensional data without reliance on distributional assumptions have been proposed to check this using measures such as maximum mean discrepancy (MMD) \cite{potapov2019pt}. Such MMD-based tests have also been proposed in \cite{colbert2022evaluating} to compare the behaviors of artificial agents to those of human players for game environments. 
While two-sample testing is suitable to determine whether the distribution of the synthetic data matches that of the training data (\textit{i.e.}, the ground truth), an augmented testing framework such as \cite{meehan2020non} can be used to determine whether the generation framework produces data that overfits to the training data. This can be particularly useful to measure and control the flexibility of data generation as per the underlying requirements of the specific problem where such synthetic paths have to be generated.

\section{Model-Free Transformation}
\label{MF}

The fundamental requirements for synthetic data generation are two-fold: (1) constructing a model that captures the multivariate data distribution, and (2) implementing a tractable sampling strategy that can generate novel in-distribution examples from the model. These two objectives can be achieved by 
constructing mappings or estimating distributions to enable efficient sampling. In particular, DGMs learn transformations between a real-world data distribution $X$ and a simpler latent distribution $Z$, such as Gaussian $N(0, I)$ from which independent and identically distributed (IID) samples can be drawn easily.
In such cases, the general mapping of probability distributions between the original space $X$ and the latent space $Z$ is:
\[
P_X \xrightarrow{f} P_Z \xrightarrow{g} P_X,
\]
where \( P_X \) is the distribution of samples in the original space, \( P_Z \) is the distribution in the latent space from which new samples can be drawn, \( f: X \rightarrow Z \) is the learned encoding (or transformation) function, and  \( g: Z \rightarrow X \) is the learned decoding (or inverse transformation) function.

The model-free (MF) method proposed in \cite{politis2013model} uses a similar principle and constructs an invertible transformation which creates a mapping $H_n$ between data $\bm{Y}$ and 
IID samples $\bm{\epsilon}$, where $\bm{Y} = (Y_1, Y_2, \ldots, Y_n$),  $\bm{\epsilon} = (\epsilon_1, \epsilon_2, \ldots, \epsilon_n)$ and $n$ is the number of points in the data. In contrast with other methods such as DGMs, the MF transformation is performed on single data instances; for example, in regression \cite{politis2015model}, time series \cite{das2021predictive} or random fields \cite{das2023model}.
In our case, the data $\bm{Y}$ is the path of a navigation agent in a video game represented by Cartesian coordinates, which has a slow-changing stochastic structure and can be represented using a locally stationary time series (LSTS) model \cite{dahlhaus2012locally}. For this type of data, the MF transformation has three steps, which we describe below for a single dimension of the multivariate time series.

In the first step of the MF transform, we calculate the cumulative distribution function (CDF) of the time series $\{Y_t \}$ at each $t = 1, 2, \ldots, n$ using
$D_{t }(y)=P\{ Y_t\leq y  \}$. In this case, the function $D_{t}$ can be obtained at each $t$ using kernel density estimation \cite{wand1994kernel} with bandwidth $b$. After we obtain $D_t$ we can then generate correlated uniform random variables $U_t$ in the range $[0,1]$ as below:
\begin{equation}
\label{MF_t1}
\ \mbox{{\bf Step 1:}} \ \ \ \  U_t =  D_t(Y_t) \ \ \mbox{for} \ t\in\{1,\ldots,n\} ~,
\end{equation}

The next step of the MF transformation is to invert the correlated uniform variables $U_t$ to obtain correlated standard normal variables $Z_t$ using:
\begin{equation}
\label{MF_t2}
\ \mbox{{\bf Step 2:}} \ \ \ \  Z_t = \Phi^{-1} (U_t)  \ \ \mbox{for} \ t \in \{1,\ldots,n\}~,
\end{equation}
where $\Phi$ denotes the CDF of the standard normal distribution, and $Z_1,\ldots, Z_n$ are correlated standard normal variables.

Finally, we calculate the $n\times n$ covariance matrix $ \Gamma_n $ from $Z_1,\ldots, Z_n$, which can be obtained using the matrix estimator proposed in {\cite{mcmurry2010banded}}.
 Assuming that $\Gamma_n $ is invertible, we can decorrelate the standard normal variables $Z_t$ as below:
\begin{equation}
\label{MF_t3}
\ \mbox{{\bf Step 3:}} \ \ \ \  \epsilon_t =   C_n^{-1} Z_t  \ \ \mbox{for} \ t \in \{1,\ldots,n\}~,
\end{equation}
where $  C_n$ is the lower triangular Cholesky factor of $ \Gamma_n $, \textit{i.e.}, 
$ \Gamma_n  =  C_n   C_n^T$ where $C_n^T$ denotes the transpose of $C_n$. It then follows that $\epsilon_1, \ldots, \epsilon_n$ are uncorrelated standard normal. Assuming that the random variables $Z_1,\ldots, Z_n$ are jointly normal it can then be inferred that $\epsilon_1, \ldots, \epsilon_n$ are IID standard normal\footnote{Joint normality of $Z_1,\ldots, Z_n$ can be established by assuming a generative model of the LSTS, for a more detailed discussion refer to \cite{das2021predictive}}. This allows the generation of IID data that can be easily sampled for our synthetic data generation task.

For $\bm{Y} = (Y_1, \ldots, Y_n)$, steps 1-3 can be then composed into the following to describe our full MF transformation:
$$
H_{n} := C_n^{-1} \circ \Phi^{-1} \circ {D_t}~,
$$
where the inverse estimators $C_n, \Phi$, and $\bm{D_t}^{-1}$ can be used to generate the full inverse MF transformation $H^{-1}_{n} : \bm{\epsilon} \mapsto \bm{Y}$. We use these key properties to generate our synthetic paths.

\section{Copula with Model-Free Transformation}
\label{copula_MF}

The model-free framework described in the previous section enables the generation of IID standard normal samples (MF residuals), which can be transformed to samples from individual dimensions of the original $p$ dimensional multivariate time series using the inverse MF transform.
We then use a copula to model the spatial correlation of the generated IID samples across dimensions \cite{nelsen2006introduction}. The copula is used to construct the multivariate distribution of the MF residuals by coupling the individual probability distributions on each dimension.

The relationship between a copula $C$ and the corresponding $p$-dimensional multivariate CDF $F_X$ is given by Sklar's theorem \cite{ruschendorf2009distributional} via:
\begin{equation}
F_X(x_1, x_2, \ldots, x_p) = C(F_1(x_1), F_2(x_2), \ldots, F_p(x_p))~,
\end{equation}
where $F_i(x_i)$ for $i \in \{1, 2, \ldots, p\}$ represents the CDFs of each component of the $p$-dimensional multivariate distribution. The corresponding equation for the copula density function $c$, the probability density functions $f_i$ of each dimension $i$ and the multivariate probability distribution function $f_X$ is given by:
\begin{equation}
\label{eq_copdense}
f_X(x_1, \ldots, x_p) = c\left( F_1(x_1), \ldots, F_p(x_p) \right) \prod_{i=1}^{p} f_i(x_i)
\end{equation}

Let the $p$-dimensional multivariate time series under consideration at time $t$ be denoted as ${\bf Y_t} = (Y_{1t}, Y_{2t}, \ldots Y_{pt})$ where $t =1,2, \ldots, n$. We use the MF estimators as described in Equations \ref{MF_t1}, \ref{MF_t2} and \ref{MF_t3} to obtain the corresponding residuals from each dimension $\bf \epsilon_{kt}$, where $k = 1, 2, \ldots, p$. Following this we estimate the $p \times p$ correlation matrix $\Sigma$ between these residuals and use them in a Gaussian copula to model the spatial dependence across the dimensions. Sampling from this copula produces new MF residuals $\bf \epsilon_{kt}^{'}$ at each time index~$t$, which can then be transformed to novel multivariate time series samples denoted as ${\bf Y_t^{'}} = (Y_{1t}^{'}, Y_{2t}^{'}, \ldots Y_{pt}^{'})$ using the inverse MF transform.
In contrast with the density $c$ which is modeled using a Gaussian copula, the marginal distributions of the residuals of each dimension (denoted by $f_i$ in Eq. \ref{eq_copdense})  are not assumed to belong to any specific distribution such as Gaussian and can be modeled nonparametrically using kernel density estimation.

In addition to the copula sampling, we introduce two other mechanisms to generate diverse synthetic paths. First, we use the following to create residuals with correlations that are different from the original copula-generated samples:
\begin{equation}
\epsilon'' = \epsilon' C_{\text{orig}}^{-1} C_{\text{target}}~,
\label{eq:eps2}
\end{equation}
where $C_{\text{orig}}$ is the Cholesky decomposition of correlation matrix $\Gamma_{\textrm{orig}}$ from the original copula-generated samples $\epsilon'$, and $C_{\text{target}}$ is the Cholesky decomposition of correlation matrix $\Gamma_{\textrm{target}}$ from the new samples $\epsilon{''}$ with specified target values of correlation $\gamma_{ij}$ for $i,j \in \{1,2,\ldots,p\}$ and $i \neq j$. 
This transformation allows for varying the correlation structure of the copula-generated MF residuals without modifying the density functions $f_i$ on each dimension $k$ and it can therefore be used to create varying  synthetic paths ${\bf Y_t{''}}$.
A second mechanism introduced for creating path diversity is to scale the residuals $\epsilon{''}$ with a factor $\lambda$ so that in the general case we sample from $N(0, \lambda^2)$ instead of $N(0, 1)$ for synthetic path generation. Overall the parameters $b$, $\Gamma_{\textrm{target}}$, and $\lambda$ enable the generation of diverse paths whose distributional and temporal fidelity to the original paths is estimated using a nonparametric hypothesis test as outlined in the next section.

\begin{figure}[tb]
    \centerline{\includegraphics[width=0.3\textwidth]{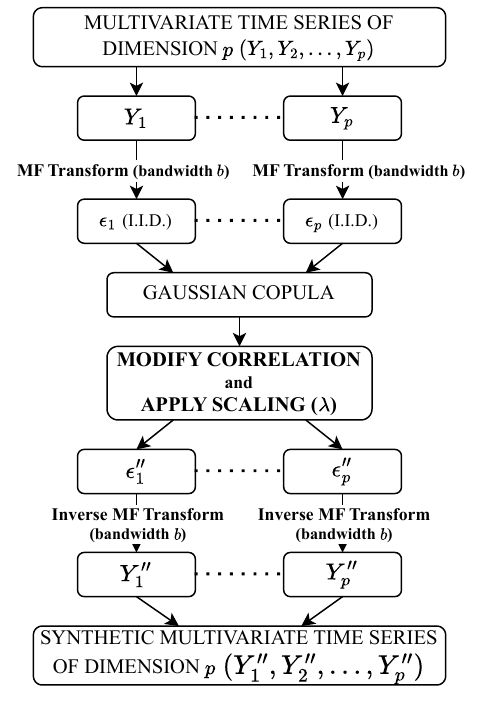}}
    \caption{Proposed flow for synthetic data generation.}
    \label{fig:copula_MF}
    \vspace{-0.5cm}
\end{figure}

Note that in the general case the spatial correlations across the $p$ dimensions may be time varying and we detect this by using a hypothesis test which is used to determine whether a path under consideration has non-stationary cross-correlation \cite{kojadinovic2024package}. If this test fails then the path is broken up into smaller segments for which the cross-correlation does not vary with time and a separate copula is used in each of the stationary segments to capture the cross-correlation. 
Fig \ref{fig:copula_MF} shows our complete synthetic path generation flow.

\section{Nonparametric testing for path validation}
\label{np_3s}

A key requirement to check the accuracy of any synthetic generation algorithm is to evaluate the validity of the generated data versus that of the training data. In addition, our proposed method (outlined in Sections \ref{MF} and \ref{copula_MF}) has the flexibility and control to generate synthetic samples of varied diversity. Therefore, beyond two-sample hypothesis tests based on distribution matching between synthetic and training samples (as in existing navigation agent hypothesis tests \cite{devlin2021navigationturingtestntt, colbert2022evaluating}), we determine when the generated samples are close to the training data (\textit{i.e.}, overfitting) versus when they are very different from the training data (\textit{i.e.}, underfitting). We accomplish this by adapting a three-sample hypothesis test as proposed in \cite{meehan2020non}.

The test determines whether synthetic samples from our generated distribution $Q$ are closer to the training set $T \sim P$ than they should be. For this purpose, we also use held out test samples $P_n \sim P$. Given $Q_m$, which denotes $m$ samples from $Q$, we conclude that $Q$ is copying $T$ when samples $x \sim Q_m$ are on average closer to $T$ than are samples $z \sim P_n$. Given a distance metric $d$ to measure proximity (in our case we choose this to be Euclidean distance which gives sufficiently good results), we use the following test statistic \cite{meehan2020non}:
\begin{equation}
\small
  \rho_T(Q_m, P_n) =  \frac{1}{nm} \sum_{x_i \in Q_m, z_j \in P_n} \mathbbm{1} \big( d(x_i, T) > d(z_j, T) \big) ~
    \label{eq:4}
\end{equation}

An ideal set of generated samples from our MF-copula algorithm will yield a statistic of $\frac{1}{2}$ when the generated samples $Q_m$ are the same distance from the training set $T$ as the test samples $P_n$. Z-scoring this statistic gives $Z_U$, where $Z_U \ll 0$ indicates overfitting, $Z_U \gg 0$ indicates underfitting, and an ideal statistic is near $0$. In general, $Z_U$ is not robust for cases where there is excessive data-copying in one region, and significant underfitting in another. To mitigate this issue, we use binning. We collect $Z_U^\pi$ in each bin $\pi$ which allows us to localize our analysis of the MF-copula fitting. Finally we compute a single statistic over these bins by taking a weighted average of $Z_U^\pi$ denoted as $C_T$ as given below in Equation (\ref{eq:5}) where $C_T \ll 0$ indicates overfitting, $C_T \gg 0$ indicates underfitting, and an ideal statistic is near $0$ \cite{meehan2020non}.
\begin{equation}
    \scalebox{1}{$\displaystyle C_T = \sum_{\pi\in\Pi}{\frac{\#\{P_n\in\pi\}}{n} Z_U^\pi}$}
    \label{eq:5}
\end{equation}

To create the samples $x \sim Q_m$ and $z \sim P_n$ for the test, we modify our $p$-dimensional time series instances as follows: For each series under consideration, we create the samples used in the test by choosing a subsequence of length $L$ on each dimension and then stacking them $p$ times. 
The dimensionality of the resulting samples is then $pL$. In this way, our method of subsequence embedding captures local temporal dependencies and adapts the original test to our time series path data. It creates point clouds from time series, which aligns with the way the original test treats images as points in high-dimensional space. We choose $L$ such that it is not too small, which would make the test be overly sensitive to small fluctuations in the data whereas very large values of $L$ would result in too few points resulting in a loss of testing power. As a practical heuristic, we set $L$ to be of the order $O(\sqrt{T})$, which provides a reasonable trade-off between these two effects. This choice is consistent with existing heuristics in time series analysis, where subsequence or window lengths are often chosen to be sublinear in $T$ to balance robustness and sample size considerations \cite{fraikin2023t}.

\section{Empirical Validation}
\label{ev}

We examine our MF-copula approach on two in-game navigation path datasets. The first dataset is a collection of samples created from AMD Schola's Tag example \cite{schola}, where three Chaser agents are pursuing a Runner agent; all agents are driven by the same neural network pre-trained using RL. The second dataset is from the Navigation Turing Test \cite{devlin2021navigationturingtestntt}, which provides path data from humans and two types of artificial neural agents: (1) the symbolic agent, which observes low-dimensional representations of the goal and game state, and (2) the hybrid agent, which receives both symbolic observations and a $32\times32$ depth buffer of game footage.

\begin{figure}[tb]
\centering
    \centerline{\includegraphics[width=0.45\textwidth]{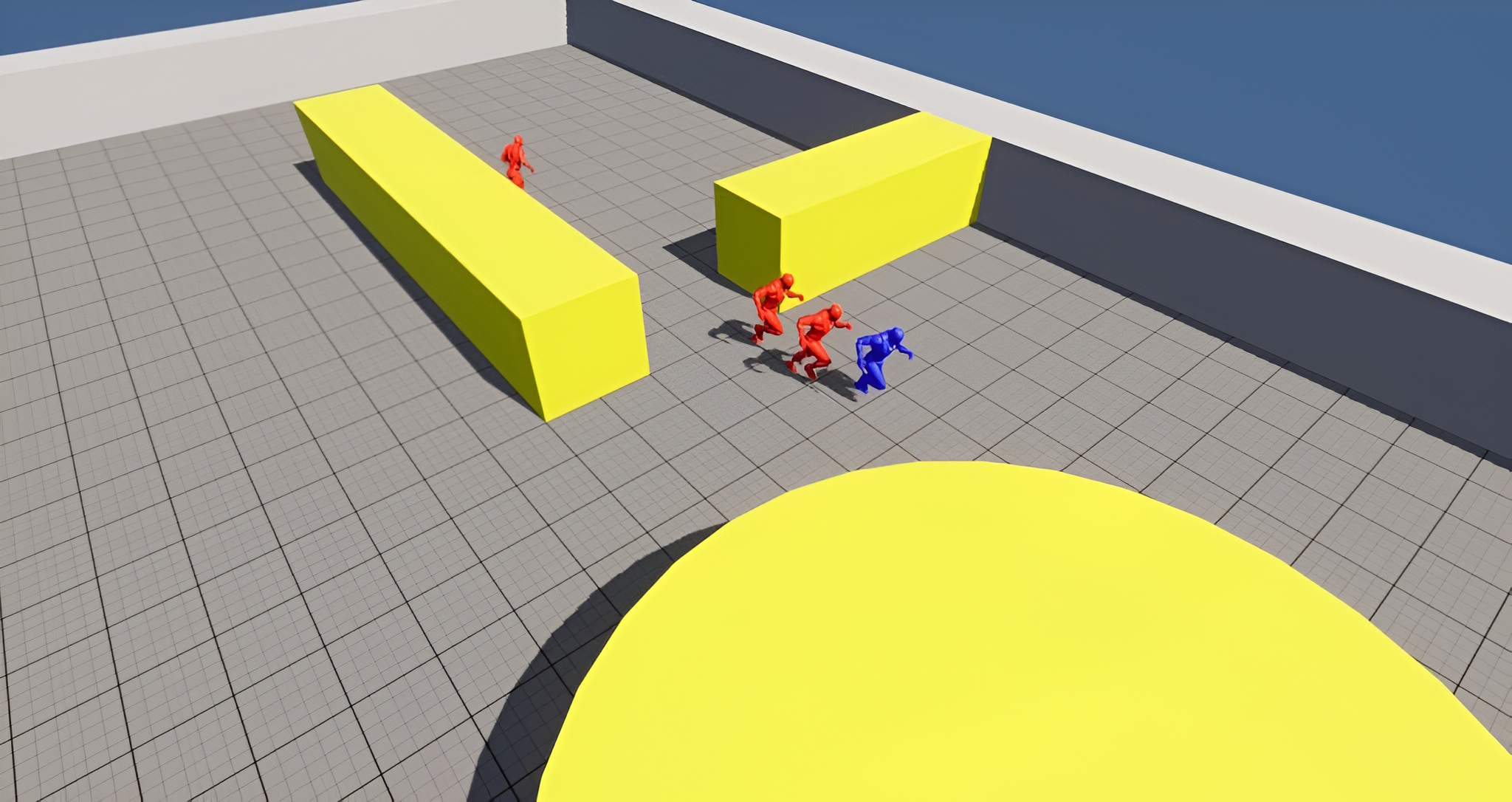}}
    \caption{An example in Schola plugin \cite{schola}. Reproduced with permission.}
    \label{fig:schola}
\end{figure}

\begin{figure}[tb]
\centering
    \centerline{\includegraphics[width=0.75\linewidth]{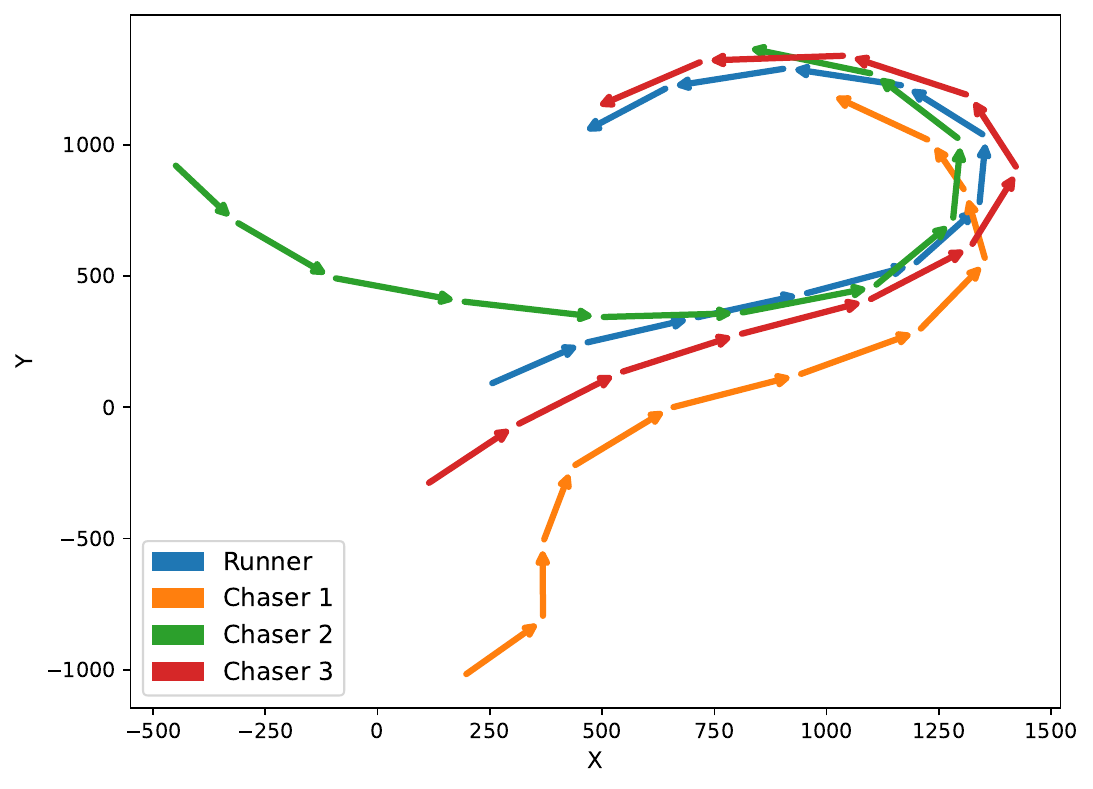}}
    \caption{Example of a scenario for the environment shown in Fig. \ref{fig:schola}. We extract 2D time series of each agents $x, y$ coordinates.}
    \label{fig:scen0}
    \vspace{-0.25cm}
\end{figure}

\textbf{Tag with Schola}. We used Schola \cite{schola} to generate path traces from the Tag  example. A sample Tag environment is shown in Fig. \ref{fig:schola}. Fig. \ref{fig:scen0} demonstrates an example of the extracted $x,y$ coordinate trajectories tracked over time.
In our synthetic data generation flow, each of these four sets of paths are used by the MF-copula algorithm to produce multiple paths. This approach prioritizes flexibility over efficiency.
When compared to an agent-ensembling approach, our agent-specific approach also boosts the interpretability of results by directly linking each synthetic trajectory to an agent, thereby allowing for straightforward comparisons.

\begin{figure}[tb]
    \centerline{\includegraphics[width=0.82\linewidth]{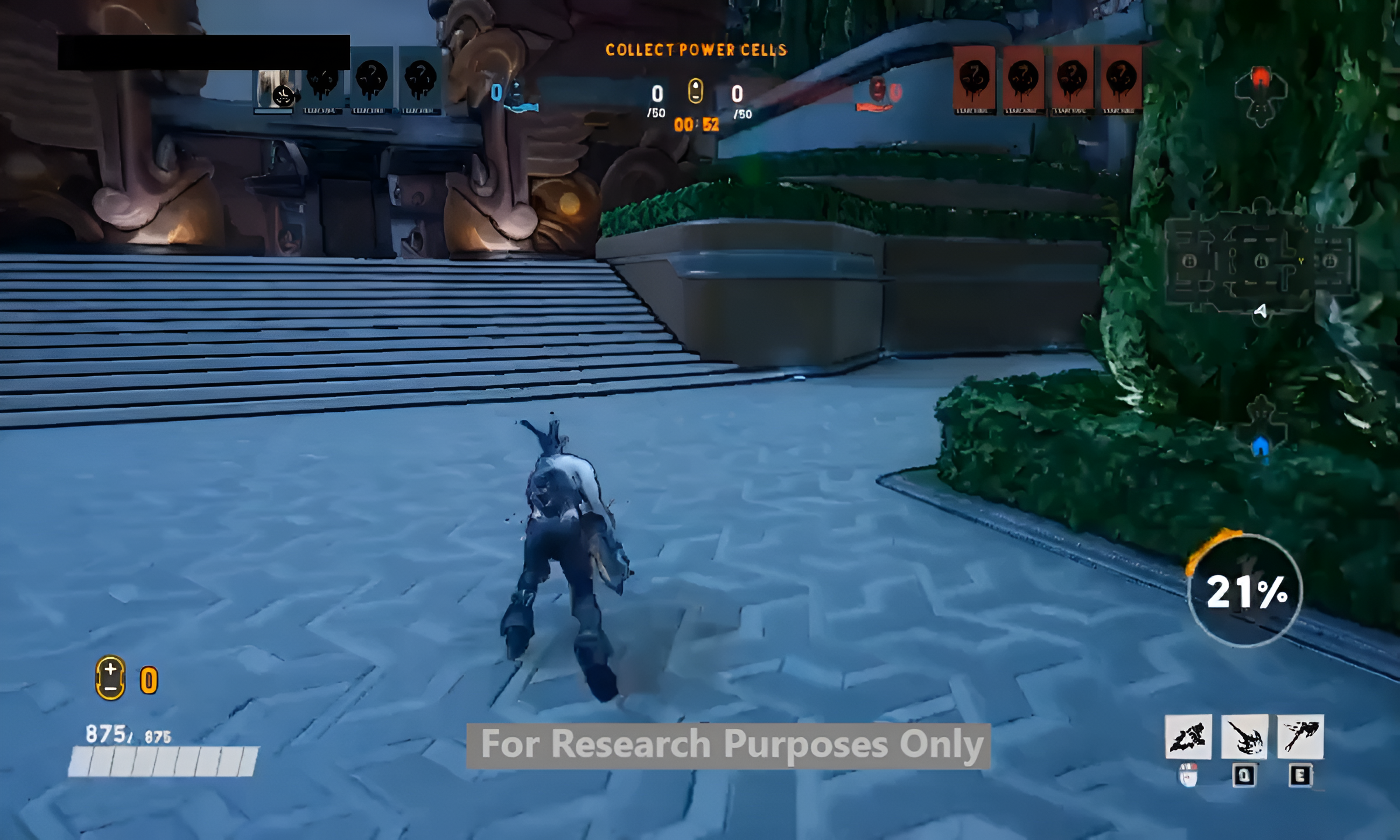}}
    \caption{A sample video frame from the NTT game data \cite{devlin2021navigationturingtestntt}. Reproduced with permission.}
    \label{fig:ntt}
\end{figure}

\begin{figure}[tb]
    \centerline{\includegraphics[width=0.8\linewidth]{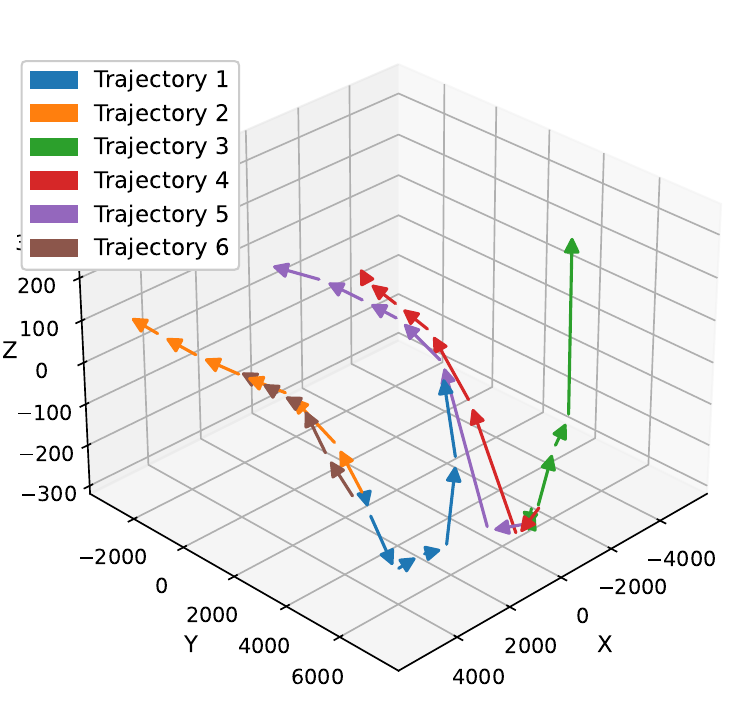}}
    \caption{Example of extracted data from NTT game data in Fig. \ref{fig:ntt} The NTT dataset contains trajectories with a variety of navigational behaviors. These trajectories have various start and end points with varying lengths and shapes.}
    \label{fig:ntt-s0}
    \vspace{-0.25cm}
\end{figure}

\textbf{Navigation Turing Test (NTT).} Devlin \textit{et al.}~\cite{devlin2021navigationturingtestntt} built the NTT dataset using an environment that is representative of a AAA video game. The 3D paths for this dataset are generated from games played by human and artificial agents; Figure \ref{fig:ntt} shows a screenshot of the game. We use the $x,y,z$ coordinates of the agent extracted at each frame, examples of which are given in Fig. \ref{fig:ntt-s0}.

% \vspace{-0.2cm}

\section{Results}
\label{res}

In this section, we demonstrate visualizations and hypothesis testing results for the synthetic data generated using the two datasets introduced in Section \ref{ev}. For synthetic realizations, we use a set of uniformly distributed correlations which are within $\pm \ 0.2$ of the values in the original correlation matrix $\Gamma_{\textrm{orig}}$ estimated from the training data. The bandwidth $b$ of our kernel density estimator used in the MF transform, and our variance scaling parameter $\lambda$ are used to control the diversity of the generated data.

Both the Tag and NTT data are characterized by nonstationary cross-correlations which are detected by applying the hypothesis test proposed in \cite{kojadinovic2024package} on the MF residuals. Based on the results, the original paths are partitioned into locally stationary sections and the MF-copula algorithm is then applied on each section. The final continuous synthetic path is generated by stitching together the results from each partition.

\begin{figure*}[!htbp]
    \centerline{\includegraphics[width=0.93\linewidth]{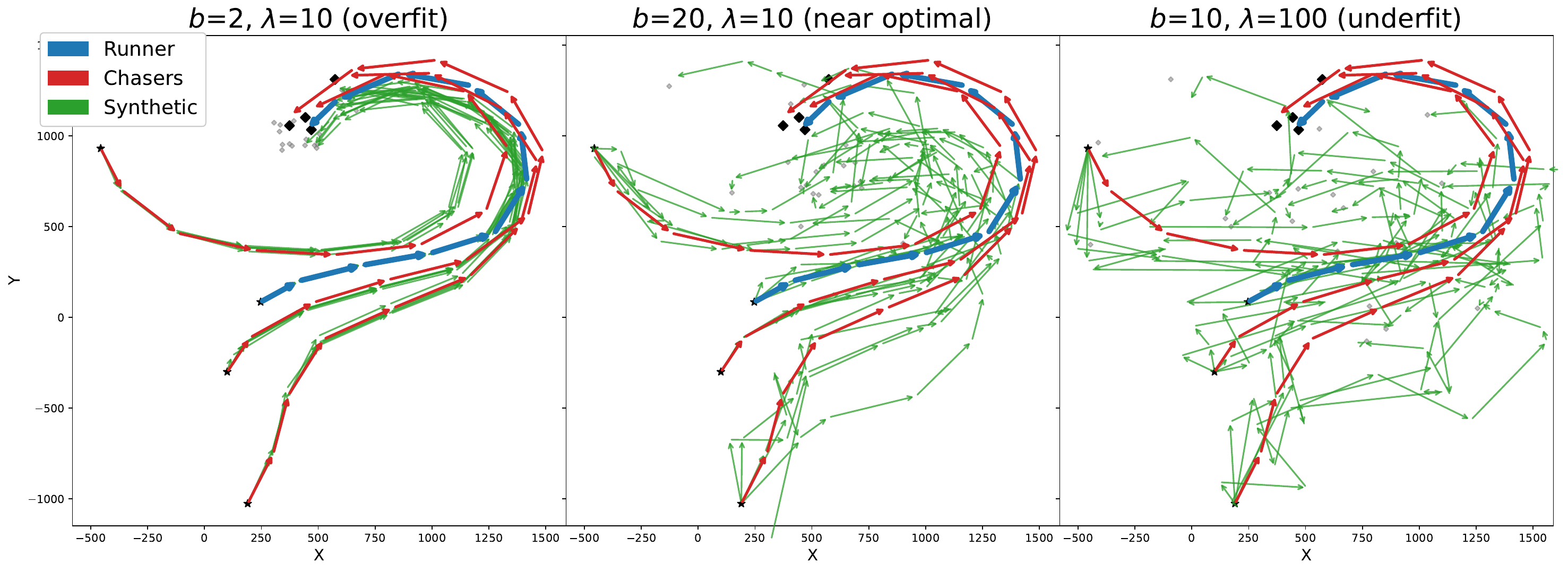}}
    \caption{Original and Synthetic paths from Schola plugin. Note that original paths are shown in red and blue, starting from stars and ending at diamonds. Synthetic paths for each case are shown in green (5 for each agent). We see that increasing $b$ or $\lambda$ increases the variance in the synthetic paths and they transition from being strongly overfitted to strongly underfitted from left to right. While the middle subplot produces most of its variance towards the end of the synthetic trajectories, the right subplot produces high variance even at the start. This is likely the reason for the increased $C_T$ value observed in Table \ref{tab:schola-experiments}.}
    \label{fig:schola-experiments}
\end{figure*}

\begin{table}[tb]
\caption{3-Sample Testing Results for One Selected Schola Path.}
\vspace{-0.6cm}
\begin{flushleft}
\end{flushleft}
\begin{center}
\begin{tabular}{|c|c|c|c|c|c|}
    \hline
    \multirow{2}{*}{\textbf{$b$}} & \multirow{2}{*}{\textbf{$\lambda$}} &\multicolumn{4}{|c|}{\textbf{{$C_T$}}} \\
    \cline{3-6}
     & & \textbf{Runner}& \textbf{Chaser 1}& \textbf{Chaser 2}& \textbf{Chaser 3} \\
    \hline
    2& 10& -2.68& -0.64 & -1.45& -1.28 \\
    20& 10& 0.34& 1.65 & 0.55& 0.27 \\
    10& 100& 1.45& 2.04 & 1.22& 1.35 \\
    \hline
\end{tabular}
\label{tab:schola-experiments}
\end{center}
\vspace{-0.65cm}
\end{table}

\begin{figure*}[!htbp]
    \centerline{\includegraphics[width=0.9\linewidth]{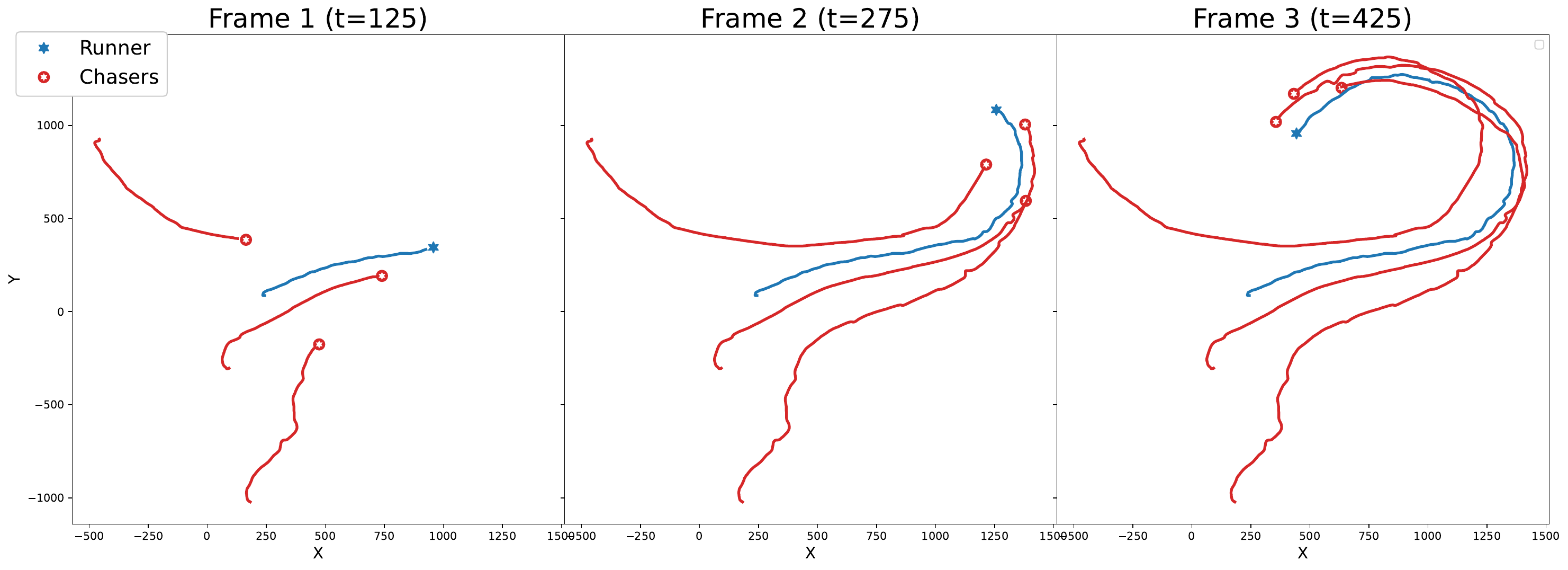}}
    \caption{Three frames from a synthetically generated episode based on the Schola Tag example. This figure demonstrates that temporal properties are preserved during our generation process. The trajectories progress over time in a  manner that matches the behavior of the paths in the original dataset.}
    \label{fig:scen0-gif}
\end{figure*}

\textbf{Schola.} To demonstrate the capabilities of our MF-copula method along with our adapted three-sample hypothesis test, we select a single path from the Schola plugin.
As per Equation \ref{eq:4}, the set $T$ consists of samples based on our subsequence embedding from the original training path, $P_n$ consists of such samples from a similar path in the dataset, and $Q_m$ consists of such samples from a synthetic path.
Our $C_T$ values are estimated as the mean over 50 synthetic realizations.
The results of this synthetic replication are shown in Table \ref{tab:schola-experiments}.
For low values of bandwidth $b$  and scaling $\lambda$, as in the first row, the corresponding $C_T$ values 
are negative for all agents. This indicates data copying or overfitting, which is also confirmed by our visual analysis as shown on the leftmost panel of Fig. \ref{fig:schola-experiments}. In the other two settings with larger $(b, \lambda)$, we obtain positive values of $C_T$ for all agents which indicate close to optimal fit or underfitting as shown in the middle two panels of Fig. \ref{fig:schola-experiments}. In general, a $C_T$ value close to 0 balances between data-copying and underfitting. To demonstrate that temporal properties are preserved by our synthetic generation, Fig. \ref{fig:scen0-gif} shows three frames from a synthetically generated episode. It can be seen that from Table \ref{tab:schola-experiments} that Chaser 1 appears to be an outlier, indicating that it may have different underlying characteristics than the other trajectories. Note that this supports our agent-specific modeling approach as one common set of parameters is unlikely to be optimal for synthetic replication from all types of agents, which are generated separately in our setup.

\begin{figure*}[!htbp]    \centerline{\includegraphics[width=0.9\linewidth]{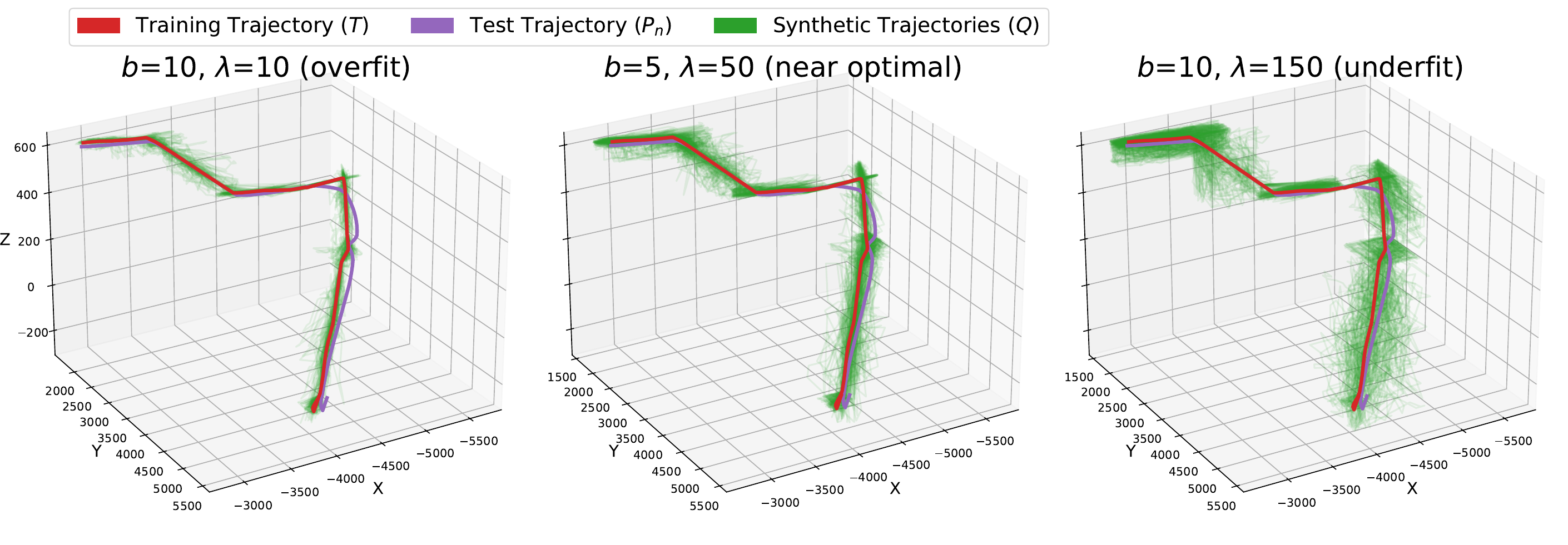}}
    \caption{50 NTT synthetic generated paths from 3 different parameter sets. 
     In this case too we see that increasing $b$ or $\lambda$ increases the variance in the synthetic paths and they transition from being strongly overfitted to strongly underfitted from left to right.}
    \label{fig:ntt-experiments}
\end{figure*}

\begin{table}[tb]
\caption{3-sample testing results for one selected NTT path}
\vspace{-0.6cm}
\begin{flushleft}
\end{flushleft}
\begin{center}
\begin{tabular}{|c|c|c|}
    \hline
    \textbf{$b$}& \textbf{$\lambda$}& \textbf{$C_T$} \\
    \hline
    10& 10& -1.60 \\
    5& 50& 0.17 \\
    10& 150& 1.45 \\
    \hline
\end{tabular}
\label{tab:ntt-experiments}
\end{center}
\vspace{-0.75cm}
\end{table}

\textbf{NTT.} The results of our synthetic replication for a single path selected
from the NTT dataset are shown in Table \ref{tab:ntt-experiments}. 
Again, as per Equation \ref{eq:4}, $T$ consists of samples based on our subsequence embedding from the original training path, $P_n$ consists of such samples from a similar path in the dataset, and $Q_m$ consists of such samples from a synthetic path. Our $C_T$ values are estimated as the mean over 50 synthetic realizations.
Similar to the Schola data, low values of bandwidth $b$ and scaling $\lambda$, as in the first row, correspond with $C_T$ values that are negative, indicating overfitting, while higher values of the yields positive $C_T$ scores that nearing optimal fit or crossing into underfitting. These trends are also supported by the path visualizations in Fig. \ref{fig:ntt-experiments}.

\textbf{Human-Like Navigation.} In addition to replicating single NTT human paths, we also use the NTT artificial agent data (\textit{i.e.}, their hybrid and symbolic agent paths). As per the results in the original NTT paper, the hybrid or symbolic paths are both distinguishable from the human-generated paths by both human and neural network-based judges \cite{devlin2021navigationturingtestntt}. We use our MF-copula generation along with the three-sample test to determine whether the synthetic human paths also have this property. For this purpose we use a \textit{distribution} of 100 human, 100 hybrid, or 100 symbolic paths instead of single paths, as in previous experiments. We generate 500 synthetic human paths by generating 5 synthetic samples from each original path.
As per Equation \ref{eq:4}, $T$ consists of samples based on our subsequence embedding from the distribution of human paths, $P_n$ consists of such samples from the distribution of either hybrid or symbolic paths, and $Q_m$ consists of such samples from the distribution of our generated human paths.
Table \ref{tab:ntt-analysis} shows the results comparing our synthetically generated human data against the hybrid and symbolic agents paths from \cite{devlin2021navigationturingtestntt}. 
Our results show that our MF-copula technique can produce paths that overfit to the original human paths, \textit{i.e.}, they are farther from the RL agent paths as shown in the first two rows of Table \ref{tab:ntt-analysis}. We can also produce high-variability trajectories, which are farther from the original human paths with respect to the artificial agent paths as shown by rows 3 and 4 of Table \ref{tab:ntt-analysis}. This showcases the flexibility of our MF-copula algorithm for producing paths of different variability with respect to the ground truth. For the case of NTT-type settings, our synthetic paths can maintain human-like dependencies with controlled variation.

\begin{table}[tb]
\centering
\caption{3 sample testing results for distributions of NTT paths}
\vspace{-0.6cm}
\begin{flushleft}
\end{flushleft}
\begin{tabular}{|c|c|c|c|}
    \hline
    \textbf{Agent}& \textbf{$b$}& \textbf{$\lambda$}& \textbf{$C_T$} \\
    \hline
    Hybrid& 10& 10& -10.05 \\
    Symbolic& 10& 10& -5.45 \\
    Hybrid& 10& 150& 10.04 \\
    Symbolic& 10& 150& 13.26 \\
    \hline
\end{tabular}
\label{tab:ntt-analysis}
\vspace{-0.5cm}
\end{table}

\section{Conclusions}
\label{conc}

In this paper, we have presented a novel statistical framework for generating synthetic navigation paths from real path traces. Our approach integrates a model-free transformation with copulas to effectively capture both temporal patterns within dimensions and complex dependencies across dimensions, without requiring large datasets or imposing restrictive parametric assumptions. Our experimental results across multiple gaming datasets demonstrate that our method successfully generates realistic synthetic paths with controlled variability. Based on a set of user controllable parameters, we can produce paths ranging from those closely matching original ground truth trajectories to those exhibiting greater diversity while maintaining plausible behavior patterns. 
Our method therefore provides a simple and effective way to control variability in synthetic generation even in the case where the amount of original data is limited, \textit{i.e.}, a single path trace.
As such, our approach offers significant practical value for game development processes, particularly in scenarios where collecting extensive human gameplay data is prohibitively expensive or time-consuming. By generating diverse yet believable synthetic paths from limited samples, our method enables more rigorous testing of navigation in games, faster iteration cycles in development, and more natural behavior patterns for non-player characters.
Our work bridges the gap between statistical time series modeling and practical game development needs, 
and effectively addresses data augmentation challenges in game applications where data scarcity is often a bottleneck. Our future work will focus on leveraging these synthetic paths for RL-based training of agents.

\section*{Acknowledgements}

From AMD, we thank Gabor Sines, Alex Cann, and Tian Liu for feedback and support.

\bibliographystyle{IEEEtran}

\bibliography{resources/srinjoy_stats}

\end{document}